%% file: paper.tex
\documentclass[sigconf,natbib]{acmart}

\usepackage{graphicx} 

\usepackage{algorithmicx}
\usepackage{algorithm}
\usepackage{algpseudocode}

\usepackage{graphicx}
\usepackage{textcomp}
\usepackage{xcolor}
\usepackage[flushleft]{threeparttable}
\usepackage{booktabs}
\usepackage{amsopn}
\usepackage{tabularx}
\usepackage{xspace}

\usepackage{amsmath,amsfonts}

\usepackage{amssymb}

\usepackage{subfig}
\usepackage{graphicx}
\usepackage{tikz}

\newcommand{\textBF}[1]{%
    \pdfliteral direct {2 Tr 0.3 w} 
     #1%
    \pdfliteral direct {0 Tr 0 w}%
}

\usepackage[backgroundcolor=yellow,textwidth=2cm]{todonotes}
\newcommand{\gk}[1]{\todo[inline,size=\small,author=GK]{#1}}

\def\mazi{Mazi\xspace}

\algnotext{EndProcedure}
\algnotext{EndFor}

\AtBeginDocument{%
  \providecommand\BibTeX{{%
    \normalfont B\kern-0.5em{\scshape i\kern-0.25em b}\kern-0.8em\TeX}}}

\setcopyright{acmcopyright}
\copyrightyear{2018}
\acmYear{2018}
\acmDOI{10.1145/1122445.1122456}

\acmConference[Woodstock '18]{Woodstock '18: ACM Symposium on Neural
  Gaze Detection}{June 03--05, 2018}{Woodstock, NY}
\acmBooktitle{Woodstock '18: ACM Symposium on Neural Gaze Detection,
  June 03--05, 2018, Woodstock, NY}
\acmPrice{15.00}
\acmISBN{978-1-4503-XXXX-X/18/06}

\begin{document}

\title{Joint Learning of Hierarchical Community Structure and Node Representations: An Unsupervised Approach}


\author{Ancy Sarah Tom}
\email{tomxx030@umn.edu}
\affiliation{%
  \institution{University of Minnesota, Twin Cities}
}
\author{Nesreen K. Ahmed}
\email{nesreen.k.ahmed@intel.com}
\affiliation{%
  \institution{Intel Labs}
}
\author{George Karypis}
\email{karypis@cs.umn.edu}
\affiliation{%
  \institution{University of Minnesota, Twin Cities}
}

\input{abstract}




\maketitle

\input{sec-intro}

\input{sec-prelim}

\input{sec-framework}

\input{sec-experiments}

\input{sec-related}

\input{sec-conclusion}

\bibliographystyle{ACM-Reference-Format}
\bibliography{paper}

\appendix
\input{sec-appendix}

\end{document}

%% file: abstract.tex
\begin{abstract}
Graph representation learning has demonstrated improved performance in tasks such as link prediction and node classification across a range of domains.
Research has shown that many natural graphs can be organized in hierarchical communities, leading to approaches that use these communities to improve the quality of node representations.
However, these approaches do not take advantage of the learned representations to also improve the quality of the discovered communities and establish an iterative and joint optimization of representation learning and community discovery.
In this work, we present \emph{\mazi}, an algorithm that jointly learns the hierarchical community structure and the node representations of the graph in an unsupervised fashion.
To account for the structure in the node representations, \emph{\mazi} generates node representations at each level of the hierarchy, and utilizes them to influence the node representations of the original graph. Further, the communities at each level are discovered by simultaneously maximizing the modularity metric and minimizing the distance between the representations of a node and its community.  Using multi-label node classification and link prediction tasks, we evaluate our method on a variety of synthetic and real-world graphs and demonstrate that \emph{\mazi} outperforms other hierarchical and non-hierarchical methods.
\end{abstract}

%% file: sec-intro.tex
\section{Introduction}
\label{sec:intro}

Representation learning in graphs is an important field, demonstrating good performance in many tasks in diverse domains, such as social network analysis, user modeling and profiling, brain modeling, and anomaly detection~\cite{hamilton2017representation}.
Graphs arising in many domains are often characterized by a hierarchical community structure~\cite{newman2006modularity, clauset2006structural, ahn2010link}, where the communities (i.e., clusters) at the lower (finer) levels of the hierarchy are better connected than the communities at the higher (coarser) levels of the hierarchy.
%
%
%
For instance, in a large company, the graph that captures the relations (edges) between the different employees (nodes) will tend to form communities at different levels of granularity. The communities at the lowest levels will be tightly connected corresponding to people that are part of the same team or project, whereas the communities at higher levels will be less connected corresponding to people that are part of the same product line or division.

In recent years, researchers have conjectured that when present, the hierarchical community structure of a graph can be used as an inductive bias in unsupervised node representation learning. 
This has led to various methods that learn node representations by taking into account a graph's hierarchical community structure.
%
%
%
\emph{HARP}~\cite{chen2018harp} advances from the coarsest level to the finest level to learn the node representations of the graph at the coarser level, and then use it as an initialization to learn the representations of the finer level graph.
%
%
%
\emph{LouvainNE}~\cite{bhowmick2020louvainne} uses a modularity-based~\cite{newman2006modularity} recursive decomposition approach to generate a hierarchy of communities. For each node, it then proceeds to generate representations for the different sub-communities that it belongs to. These representations are subsequently aggregated in a weighted fashion to form the final node representation, wherein the weights progressively decrease with coarser levels in the hierarchy.
\emph{SpaceNE}~\cite{long2019hierarchicalsubspace} constructs sub-spaces within the feature space  to represent different levels of the hierarchical community structure, and learns node representations that preserves proximity between vertices as well as similarities within communities and across communities.

Further, in recent times, certain GNN-based approaches~\cite{li2020graph,zhong2020hierarchical} have also been proposed which exploit the hierarchical community structure while learning node representations. However, these methods use supervised learning and require more information to achieve good results. 
%
%
%

Though all of the above methods are able to produce better representations by taking into account the hierarchical community structure, the information flow is unidirectional---from the hierarchical communities to the node representations. We postulate that the quality of the node representations can be improved if we allow information to also flow in the other direction---from the node representations to hierarchical communities---which can be used to improve the discovered hierarchical communities. Moreover, this allows for an iterative and joint optimization of both the hierarchical community structure and the representation of the nodes.

We present \emph{\mazi}\footnote{\mazi~is Greek for together.}, an algorithm that performs a joint unsupervised learning of the hierarchical community structure of a graph and the representations of its nodes. The key difference between \emph{\mazi} and prior methods is that the community structure and the node representations help improve each other.
%
%
%
%
\emph{\mazi} estimates node representations that are designed to encode both local information and information about the graph’s hierarchical community structure. By taking into account local information, the estimated representations of nodes that are topologically close will be similar. By taking into account the hierarchical community structure, the estimated representations of nodes that belong to the same community will be similar and that similarity will progressively decrease for nodes that are together only in progressively coarser-level communities.

\emph{\mazi} forms successively smaller graphs by coarsening the original graph using the hierarchical community structure such that the communities at different levels represent nodes in the coarsened graphs. Then, iterating over all levels, \emph{\mazi} learns node representations at each level by maximizing the proximity of the representation of a node to that of its adjacent nodes while also drawing it closer to the representation of its community.
Furthermore, at each level, \emph{\mazi} learns the communities by taking advantage both of the graph topology and the node representations.
%
This is done by simultaneously maximizing the \emph{modularity} of the communities, maximizing the affinity among the representations of near-by nodes by using a Skip-gram~\cite{mikolov2013efficient} objective, and minimizing the distance between the representations that correspond to a node and its parent in the next-level coarser graph.

We evaluate \emph{\mazi} on the node classification and the link prediction tasks on synthetic and real-world graphs. 
Our experiments demonstrate that \emph{\mazi} achieves an average gain of $215.5\%$ and $9.3\%$ over competing approaches on the link prediction and node classification tasks, respectively.

The contributions of our paper are the following:
\begin{enumerate}
\item  We develop an unsupervised approach to simultaneously organize a graph into hierarchical communities and to learn node representations that account for that hierarchical community structure. We achieve this by introducing and jointly optimizing an objective function that contains (i) modularity- and skip-gram-based terms for each level of the hierarchy and (ii) inter-level node-representation consistency terms.

\item We present a flexible synthetic generator for graphs that contain hierarchically structured communities and community-derived node properties. We use this generator to study the effectiveness of different 
node representation learning algorithms.

\item  We show that our method learns node representations that outperform competing approaches 
on synthetic and real-world datasets for the node classification and link prediction tasks. 

\end{enumerate}

%% file: sec-prelim.tex
\section{Definitions and Notation}
\label{sec:prelim}

\begin{table}[t!]
\caption{Summary of notation.}
\vspace{-3mm}
\renewcommand{\arraystretch}{1.15} 
\scalebox{1.0}{
\centering 
\fontsize{8}{9.5}\selectfont
\setlength{\tabcolsep}{5pt} 
\label{tab:notation}
\hspace*{-2.5mm}
\begin{tabularx}{1.0\linewidth}{@{}r X@{}} 
\toprule
\textbf{Notation} & \textbf{Description} \\
\midrule
$l$     & A level in the hierarchical structure. \\
$L$     & The number of levels in the hierarchical communities. \\
$G^l$   & The graph $G^l=(V^l, E^l, W^l)$ at level $l$, where $V^l$ is the set of nodes, $E^l$ is the set of edges, and $W^l$ stores the edge weights. \\
 $v_i$       & A vertex in $G$. \\
 $\text{deg}(v_i)$  & The degree of node $v_i$. \\
 $X$ & The node representations of $G$ \\
 $\mathbb{C}$ & A community decomposition of $G$. \\
 $H$ & The community membership indicator vector of $G$. \\
 $C_i$ & A community in $C$. \\
 $\text{deg}_{int}(C_i)$        & The internal degree of community ${C}_i$. \\
 $\text{deg}_{ext}(C_i)$        & The external degree of community ${C}_i$. \\
 $\text{deg}(C_i)$              & The overall degree of community ${C}_i$. \\
 $ID$   & An array containing the vertex internal degrees. \\
 $ED$   & An array containing the vertex external degrees. \\
 $Q$    & The modularity of $G$ for a given $\mathbb{C}$ (cf. Eqn.~\ref{modularity}).\\
 $X^l$  & The node representations at level $l$. \\ 
 $H^l$  & The community structure at level $l$. \\
 $d$    & The dimension of $X^l$, where $l \in 1,\ldots,L$. \\
 $ne^l$ & The number of epochs at level $l$. \\
 $lr^l$ & The learning rate at level $l$. \\
 $k$    & The context size extracted from walks. \\
 $wl$   & The length of random-walk. \\
 $r$    & The number of walks per node. \\
 $\alpha$ & The weight of the contribution of node neighborhood to the  overall loss. \\
 $\beta$ & The weight of the contribution of proximity to a node's community to the overall loss. \\
 $\gamma$ & The weight of the contribution of $Q$ to the overall loss. \\
\bottomrule
\end{tabularx}}
\end{table}

Let $G=(V,E)$ be an undirected graph where $V$ is its set of $n$ nodes and $E$ is its set of $m$ edges. 
Let $\mathbf{X} \in \mathbb{R}^{n \times d}$ store the representation vector $x_i$ at the $i$th row for $v_i \in V$.

A \emph{community} refers to a group of nodes that are better connected with each other than with the rest of the nodes in the graph. 
A graph is said to have a \emph{community structure}, if it can be decomposed into communities.
In many natural graphs, communities often exist at different levels of granularity. At the upper (coarser) levels, there is a small number of large communities, whereas at the lower (finer) levels, there is a large number of small communities. In general, the communities at the coarser levels are less well-connected than the finer level communities. When the communities at different levels of granularity form a hierarchy, that is, a community at a particular level is fully contained within a community at the next level up, then we will say that the graph has a \emph{hierarchical community structure}.


%
%

Let $\mathbb{C} = \{C_0, \ldots, C_{k-1}\}$, with $V = \cup_i C_i$ and $C_i\cap C_j = \emptyset$ for $0\le i, j < k$ be a $k$-way \emph{community decomposition} of $G$ with $C_i$ indicating its $i$th community. 
Let $H$ be the \emph{community membership} indicator vector where $0\le H[v_i] < k$ indicates $v_i$'s community.
Given a $k$-way community decomposition $\mathbb{C}$ of $G^l=(V^l, E^l)$, its \emph{coarsened} graph $G^{l+1} = (V^{l+1}, E^{l+1})$ is obtained by creating $k$ vertices---one for each community in $\mathbb{C}$---and adding an edge $(v_i, v_j)\in E^{l+1}$ if there are edges $(u_p, u_q)\in E^l$ such that $u_p\in C_i$ and $u_q\in C_j$. The weight of the $(v_i, v_j)$ edge is set equal to the sum of the weights of all such $(u_p, u_q)$ edges in $E^l$. In addition, each $v_i\in V^{l+1}$ is referred to as the parent node to all $u\in C_i$.

Given $\mathbb{C}$, the \emph{modularity} of $G$ is defined as
\begin{equation}
    \label{modularity}
    Q = \frac{1}{2m}\Bigg(\sum_{C_i \in \mathbb{C}}\bigg(\text{deg}_{int}(C_i) - \frac{\text{deg}(C_i)^2}{2m}\bigg)\Bigg).
\end{equation}
Here, $\text{deg}_{int}(C_i)$ is the number of edges that connect nodes in $C_i$ to other nodes in $C_i$ and $\text{deg}(C_i)$ is the sum of all node degrees in $C_i$. 
Further, let $\text{deg}_{ext}(C_i)$ be the number of edges that connect $C_i$ to nodes in other communities.
$Q$ measures the difference between the actual number of edges within $C_i$ and the expected number of edges within $C_i$, aggregated over all $C_i \in \mathbb{C}$. 
$Q$ ranges from $-0.5$, when all the edges in $G$ are between $C_i$ and $C_j$, where $i \neq j$, and approaches $1.0$ if all the edges are within any $C_i$ and $k$ is large.

Let the hierarchical community structure of $G$, with $L$ levels, be represented by a sequence of successively coarsened graphs, denoted by $G, G^2, \cdots , G^L$, such that $|V| > |V^2| > \cdots > |V^L|$, wherein at each $l \in L$, the communities in $G^l$ are collapsed to form the nodes in $G^{l+1}$. 
Every $v^l_i \in V^l$ is collapsed to a  single parent node, $v^{l+1}_j$, in the next level coarser graph, $G^{l+1}$. 
Let us denote a model that takes the hierarchical community structure into account as hierarchical models and those that do not as flat models. Finally, we summarize all the notations in Table~\ref{tab:notation}.

%% file: sec-framework.tex
\section{\emph{\mazi}}
\label{sec:proposed}
Given a graph $G$, \emph{\mazi} seeks to jointly learn its node representations and its hierarchical community structure organized in $L$ levels.
%
%
\emph{\mazi} coarsens the graphs at all levels of the hierarchy and learns representations for all nodes. At any given level, the node representation is learned such that it is similar to those of the nodes in its neighborhood, to its community and to the nodes it serves as a community to. This ensures the node representations at all levels align with the hierarchical community structure. Further, the communities at all levels are learned by utilizing node representations along with the graph topology. 
\emph{\mazi} utilizes Skip-gram to model the similarity in the representations of a node and its neighbors. To model the similarity in the representations of node and its associated community, \emph{\mazi} minimizes the distance between the two representations.
Finally, to learn the communities, \emph{\mazi} maximizes the \emph{modularity} metric along with the above objectives.

\begin{figure}
  \includegraphics[width=1.1\columnwidth, angle=270,origin=c,trim={0 0 0 0}]{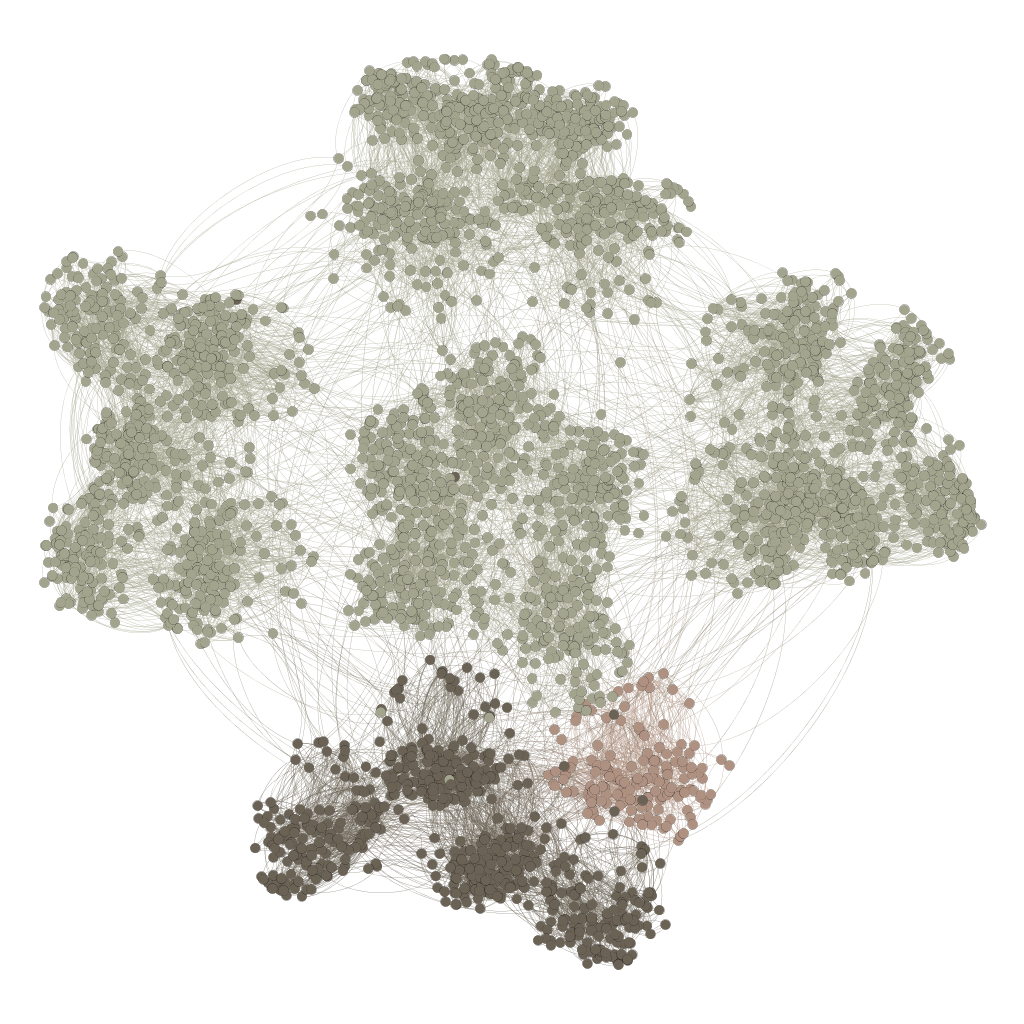}
\caption{A visualization of a synthetic $3$K-node graph with a hierarchical community structure. The graph is generated by the proposed synthetic graph generator in Section~\ref{sec:exp}. 
The important parameters
include \texttt{common-ratio} of $3.0$, a branching factor of $5$, except at the finest level, which is formed with a branching factor of $30$ and a maximum degree for each node equal to $7.5$. A community in level $3$ is colored blue and a sub-community, in level $2$, within that is represented in yellow.}
  \label{fig:hier_graph_layout}
  \vspace{-4mm}
\end{figure}

Figure~\ref{fig:hier_graph_layout} illustrates a graph with a hierarchical community structure. From the figure, we see that the original graph (level $1$ in the hierarchical structure) contains $5$ large communities (level $3$) in its coarsest level, each of which can be further split into $5$ sub-communities (level $2$).
A community in level $3$ is represented in \emph{blue} and one of its $5$ sub-communities is colored \emph{yellow}. \emph{\mazi} learns the representation of a node belonging to the \emph{yellow} community such that it will be similar to other nodes in that community over others. Furthermore, it will also be similar in representation to the nodes in the \emph{blue} community, although this similarity value will be progressively lower as compared to that of the nodes in the \emph{yellow} community. 

\subsection{Objective Function}
%
\emph{\mazi} defines the objective function used for learning node representations using three major components. 
First, at each level, for each node, \emph{\mazi} maximizes the proximity of its representation to the representation of the nodes belonging to its neighborhood using the Skip-gram objective. Second, iterating over all levels, the proximity of the representation of a node to that of its direct lineage in the embedding space is maximized. 
Third, the communities at each level are discovered and refined by maximizing the \emph{modularity} metric.

\paragraph{Modeling node proximity to its neighborhood.}
As previously studied, see~\cite{grover2016node2vec}, to capture the neighbourhood of a node in the representations, we seek to maximize the log-likelihood of observing the neighbors of a node conditioned on its representation using the Skip-gram model with negative sampling.
Utilizing the concept of sequence-based representations, neighboring nodes of a node $v_i$, represented by $N(v_i)$, are sampled to form its context. 
Let the negative sampling distribution of $v_i$ be denoted by $P_n$ and the number of negative samples considered for training the loss be denoted by $R$. 
We use $L_{nbr\_pos}$ and $L_{nbr\_neg}$ to denote the loss of $v_i$ to its neighbors and to its negative samples, respectively. Using the above, we define 
\begin{subequations}
\begin{align}
%
L_{nbr\_pos} &= \frac{1}{|N(v_i)|}\sum_{v_j\in{N(v_i)}}\log \sigma(x_i^\top x_j), \label{pos_loss}\\
L_{nbr\_neg} &= ~R \cdot E_{v_n \sim P_n(v_i)}\log (1-\sigma(x_i^\top x_j)). \label{neg_loss}
\end{align}
\end{subequations}

Taken together, we model the neighbourhood proximity of $v_i$ as: 
\begin{equation}
\label{loss_nbr}
\begin{split}
L_{nbr} &= L_{nbr\_pos}~+~L_{nbr\_neg}. 
\end{split}
\end{equation}

\paragraph{Modeling node proximity to its Community.}
In many domains, nodes belonging to a community tend to be functionally similar to each other in comparison to nodes lying outside the community~\cite{clauset2008hierarchical}. 
As a consequence, we expect the representation of a node to be similar to the representation of its lineage in the hierarchy.
%
Consider a level, $l$, in the hierarchical community structure of $G$. 
At $l$, for $v^l_i$, with representation $x_i^l$, we let the representation of its associated community (parent-node) in the next level coarser graph, $G^{l+1}$, be denoted by $x_{H^l(v^l_i)}^{l+1}$. 
%
To model the relationship between $v^l_i$ and $H^l(v^l_i)$, we use:
\vspace{-1mm}
\begin{equation}
\label{comm_level}
    L^l_{comm}~=~\log \sigma\big({x_i^{l}}^\top  {x_{H^l(v_i^l)}^{l+1}}\big).
\end{equation}

As we iterate over the levels in the hierarchy of the graph, we bring together nodes in each level closer to its parent node in the next-level coarser graph in the embedding space. Consequently, the representation of a node is influenced by the communities the node belongs to at different levels.
 
\paragraph{Jointly Learning the Hierarchical Community Structure and Node Representations.}
Typically, community detection algorithms utilize the topological structure of a graph to discover communities. However, we may also take advantage of the information contained within the node representations while forming the communities at each level in the hierarchy. \emph{\mazi} discovers the communities in the graph by jointly maximizing the \emph{modularity} metric, described in Equation~\ref{modularity}, at each level and minimizing the distance between the representations of a node and its community in the next level coarser graph. The communities that we learn at each level, thus, better align with the structural and the functional components of the graph at that level.
%
%
At each level in the hierarchical community structure, we use Equation~\ref{loss_nbr} and Equation~\ref{comm_level} to model and learn the node representations. 

Consequently, putting all the components together, we get the following coupled objective function:
\begin{equation}
\label{eqn_dbl_x}
    \begin{split}
       {\max_\theta}
       \sum_{l=1}^{L} \Bigg(&\frac{1}{|V^l|}\Big(L^l_{nbr\_pos} +\alpha^lL^l_{nbr\_neg}+\beta^lL^l_{comm}\Big)+\gamma^lQ^l\Bigg), \\
       &\theta = x_i^l, H^l,~i \in 1 \dots |V|^l~\forall l \in 1 \dots L.
    \end{split}
\end{equation}
\vspace{-3mm}
\noindent
 
Since the order of the three terms that contribute to the overall objective is different, the terms are normalized with its respective order of contribution. 
Further, $\alpha^l$, $\beta^l$ and $\gamma^l$ serve as regularization parameters and are added to Sub-equations~(\ref{neg_loss}),~(\ref{comm_level}) and~(\ref{modularity}) in the overall objective for each level $l$, respectively. 

\subsection{Algorithm}
An initial hierarchical community structure of the graph at level $1$, denoted by $G^1 = (V^1, E^1, W^1)$,  is constructed and node representations are computed for all the levels in the hierarchy. Then, using an alternating optimization approach in a level-by-level fashion, the objective, defined previously, is optimized. 
%
The optimization updates step through the levels from the finest level graph to the coarsest level graph (Forward Optimization)
and then from the coarsest level graph to the finest level graph (Backward Optimization) in multiple iterations.
This enables the node representations at each level to align itself to its direct lineage in the embedding space, additionally refining the community structure by the information contained within this space. 
An outline of the overall algorithm can be found in Algorithm~\ref{mazi}. 

\input{alg-mazi}

\vspace{-1mm}
\paragraph{Initializing the Hierarchical Community Structure and Node Representations.}
A hierarchical community structure with $L$ levels and their associated community membership vectors for $G$ is initialized by successively employing existing community detection algorithms, such as \emph{Metis}~\cite{karypis1995multilevel} at each level $l \in L$.
The node representations at the finest level of the graph, denoted by $X^1$, are initialized by using existing representation learning methods such as \emph{node2vec, DeepWalk}~\cite{grover2016node2vec, perozzi2014deepwalk}. 
Node representations of coarser level graphs are then initialized by computing the average of the representations of nodes that belong to a community in the previous level finer graph, $G^{l-1}$. 

\vspace{-2mm}
\paragraph{Optimization Strategy.}
At each level, \emph{\mazi} utilizes an alternating optimization (AO) approach to optimize its objective function. \emph{\mazi} performs AO in a level-by-level fashion, by fixing variables belonging to all the levels except one, say denoted by $l$, and optimizing the variables associated with that level.
At $l$, the community membership vector, $H^l$, is held fixed and the node representations, $X^l$, is updated. Then, $X^l$ is fixed, and $H^l$ is updated. Let us denote the node representation update as the $X^l$ sub-problem, and the community membership update as the $H^l$ sub-problem for further reference.

\paragraph{Node Representation Learning and Community Structure Refinement.}
At each level $l$, \emph{\mazi} computes the gradient updates for the $X^l$ sub-problem. By holding $H^l$ fixed, \emph{\mazi} updates $x^l_i$ to be closer to
the representation of $v_j \in N(v^l_i)$, $x^l_j$, and its parent node, $x^{l+1}_{H^l(v^l_i)}$
(see Equation~\ref{loss_nbr} and~\ref{comm_level}). 
%
The $H^l$ sub-problem is then optimized using the updated $X^l$ at $l$. To maximize the \emph{modularity} objective, \emph{\mazi} utilizes an efficient move-based approach. 
%
From Equation~\ref{modularity}, we note that $Q^l$ can be determined by computing $\text{deg}_{int}(C^l_i)$ and $\text{deg}_{ext}(C^l_i)$, where $C^l_i \in C$, and applying the above equation. Therefore, to move $v^l_i$ from $C^l_a$ to $C^l_b$, instead of computing the contribution from each community to the value of \emph{modularity}, \emph{\mazi} only modifies the internal and the external degrees of $C^l_a$ and $C^l_b$ 
by computing how the contribution of $v^l_i$ to $C^l_a$ and $C^l_b$ changes. 
%
The new community assignment of $v^l_i$ is determined such that it maximizes $Q^l$ and minimizes the distance 
between $x^l_i$ and $x^{l+1}_{H^l(v_i)}$.
This process is repeated for all nodes for a fixed number of iterations or until no moves lead to a better solution. This is returned as the optimized solution for the $H^l$ sub-problem.

After alternatively solving for the sub-problems $X^l$ and $H^l$ at level $l$, \emph{\mazi} optimizes level $l+1$. These steps proceed up the hierarchy in this fashion until it reaches level $L-1$. Starting at $L-1$, the sub-problems $X^{L-1}$ and $H^{L-1}$ is optimized in the backward direction level-by-level using the updated representations, that is, $l = L-1, L-2, \dots, 1$. By performing the optimization in the backward direction such as above, the node representations at the finer levels of the hierarchy are influenced by the updated representations at the coarser levels. 
After $W$ such iterations, the refined node representations and community membership vectors for all levels are returned as the result of the algorithm.

%% file: alg-mazi.tex
\begin{algorithm}
\small
\caption{\emph{\mazi}: Joint Unsupervised Learning of Node Embedding and Hierarchical Community Structure.}\label{mazi}
\begin{flushleft}
\textbf{INPUT:} Undirected Graph $G^1=(V^1, E^1, W^1)$
\\
\textbf{OUTPUT:} Node embedding $X^l$ and hierarchical community structure $H^l$, $\forall~l \in [1 \dots L]$  
\end{flushleft}
\begin{algorithmic}[1]
\Procedure{\mazi}{}
\State Set hyper-params $k$, $wl$, $r$, $L$, $lr^1 \dots lr^L$, $ne^1 \dots ne^L$, $W$, $d$, $\alpha$, $\beta$, $\gamma$.
\State $G^l, X^l, H^l \gets \textsc{InitGXH}(G^1, X^1, H^1, L)$ , $\forall~l \in [1 \dots L]$
\For{$w\gets 1, W$}
    \For{$l\gets 1, L-1$}
        \State $X^l, H^l \gets \textsc{UpdateXH}$\parbox[t]{.9\linewidth}{%
        $(G^l, X^l, H^l \forall~l \in [1 \dots L], l)$} \label{forward_opt}
    \EndFor \Comment {Forward Optimization: Fine to Coarse.}
    \For{$l\gets L-1, 1$}
        \State $X^l, H^l \gets \textsc{UpdateXH}$\parbox[t]{.9\linewidth}{%
        $(G^l, X^l, H^l \forall~l \in [1 \dots L], l)$} \label{backward_opt}
    \EndFor \Comment {Backward Optimization: Coarse to Fine.}
\EndFor\label{maziendfor}
\EndProcedure
\vspace{-2mm}
\\\hrulefill
\Procedure{InitGXH}{$G^1$, $X^1$, $H^1$, $L$}
\\
\Comment{\parbox[t]{.9\linewidth}{%
Initialize node representations and hierarchical community structure.}}
\label{init_ghx}
\For{$l\gets 1, L-1$}
    \ForAll {$v^l_i \in V^l$}
        \State Merge $v^l_i$ to $v^{l+1}_h$, where $v^l_i \in V^l, h \gets H^l[v^l_i]$.
    \EndFor
    \ForAll {$(v_i, v_j) \in E^l$}
        \State Collapse $(v^{l+1}_{H(v_i)}, v^{l+1}_{H(v_j)})$ to $E^{l+1}$. 
        \State $w^{l+1}_{(v^{l+1}_{H(v_i)}, v^{l+1}_{H(v_j)})} += w^{l}_{(v_i^{l}, v_j^{l})}$
    \EndFor
    \ForAll{$v_i^l \in V^l$}
        \State $x^{l+1}_{v^{l+1}_{H(v_i)}} += x^l_{v^l_i}$
    \EndFor
    \State Generate $H^{l+1}$ for $G^{l+1}(V^{l+1}, E^{l+1}, W^{l+1})$.
\EndFor
\State \textbf{return} $G^l=(V^l, E^l, W^l), X^l, H^l~\forall~l \in [1 \dots L]$
\EndProcedure
\vspace{-2mm}
\\\hrulefill
\Procedure{UpdateXH}{$G^l$, $X^l$, $H^l$ $\forall~l \in [1 \dots L]$, $l$}
\\
\Comment{\parbox[t]{.9\linewidth}{%
Update node representations and clustering solution at level $l$.}
}
\ForAll {$v_i \in V^l$}
    \For{$walk\_num\gets 1, r$} \label{update_x_begin}
        \State $train_{v_i} \leftarrow$ \texttt{RandomWalker}$(G^l, k, wl)$
        \State $X^l \gets \textsc{UpdateX}$\parbox[t]{.4\linewidth}{%
        $(X^{l-1}, H^{l-1}, X^{l}, H^{l}, X^{l+1}, \newline
        H^{l+1}, train_{v_i}, ne^l, lr^l,
        \alpha, \beta)$} 
    \EndFor \label{update_x_end}
\EndFor
\State $H^l \gets $ $\textsc{UpdateH}$\parbox[t]{.4\linewidth}{%
$(G^l, H^{l}, X^{l}, X^{l+1})$}
\State \textbf{return} $X^l, H^l$
\EndProcedure
\vspace{-2mm}
\\\hrulefill
\Procedure{UpdateH}{$G^l$, $H^l$, $X^{l}$, $X^{l+1}$}
\\
\Comment{\parbox[t]{.9\linewidth}{%
Update clustering solution.}
}
\ForAll{$v_i \in V^l$}
    \State Compute~$degrees[k]~\forall~k\gets1,|H^l|$.
    \State $deg(v_i)=$~\texttt{sum}$(degrees)$
    \For{$k\gets 1, |H^l|$}
        \State $obj[k] \gets$ $\texttt{MoveTo}$\parbox[t]{.4\linewidth}{%
        $(v_i, k, H^l(v_i), degrees, 
        \newline ID, ED, deg(v_i), X^l, X^{l+1})$}
        \State $h_{max} \gets argmax(obj)$
        \State $H^l(v_i) \gets h_{max}$
        \State Modify $ID$, $ED$.
    \EndFor
\EndFor
\State \textbf{return} $H^l$
\EndProcedure
\end{algorithmic}
\end{algorithm}

%% file: sec-experiments.tex
\section{Experiments}
\label{sec:exp}

In order to evaluate the proposed algorithm, \emph{\mazi}, in Section~\ref{sec:proposed}, we design synthetic as well as real-world experiments. We test \emph{\mazi} on two major tasks: (1) Node classification, and (2) Link prediction. 
We compare \emph{\mazi} against the below state-of-the-art baseline methods: 
\begin{itemize}
\item{\emph{node2Vec}~\cite{grover2016node2vec}: 
\emph{node2vec} uses second order random walks to capture the neighborhood of a node and optimizes its model using skip-gram with negative-sampling.}
\item{\emph{HARP}~\cite{chen2018harp}: \emph{HARP} coarsens the graph into multiple levels by collapsing edges (chosen using heavy-edge matching) and star-like structures at each level. Then, from the coarsest level to the finest level, using existing methods, such as \emph{node2vec}, node representations of the coarser level graph are generated and used as an initialization to learn the representations at the finer level graph.}
\item{\emph{LouvainNE}~\cite{bhowmick2020louvainne}: For any input graph, \emph{LouvainNE} recursively generates the sub-communities within each community in a top down fashion. For all the different sub-communities that a node belongs to, the $d$-dimensional representations are generated either randomly or using one of the existing non-hierarchical models, referred to as the \emph{stochastic} variant and the \emph{standard} variant of the algorithm respectively. These representations are subsequently aggregated in a weighted fashion to form the final node representation.}
\item{\emph{ComE}~\cite{cavallari2017learning}: \emph{ComE} jointly learns communities and node representations of a graph by modeling the community and node representations using a gaussian mixture formulation.}
\item{Variations of the above mentioned models.}
\end{itemize}


In addition, for any undirected graph, we extract the induced subgraph formed by all the vertices in the largest connected component of the graph. This pre-processing step ensures that the graphs constructed in the coarser levels in the hierarchy will remain connected. 

\subsection{Datasets}
\subsubsection*{Real World Graphs}
We evaluate the proposed algorithm on three real world networks: \emph{BlogCatalog}, \emph{CS-CoAuthor}, and \emph{DBLP}. \emph{BlogCatalog} is a social network illustrating connections between bloggers while \emph{CS-CoAuthor} and \emph{DBLP} are co-authorship networks. More information about each dataset is detailed in Table~\ref{graph_stats}. For each graph, the total number of levels in the hierarchical community structure is set equal to $4$, thereby including $2$ levels of coarsened graphs. 
The number of communities in each subsequent level is generated using $\sqrt{n}$, where, $n$ is the number of nodes in the graph in the current level.

\subsubsection*{Synthetic Graphs}
We design a novel synthetic graph generator that is capable of generating graphs with a hierarchical community structure and real-world structural properties (e.g., average degree, degree distribution, number of edges a node forms with other communities in the upper levels, etc). Figure~\ref{fig:hier_graph_layout} shows the visualization of a $3$K node graph generated with the proposed generator. We discuss the details of the proposed generator in the Appendix~\ref{sec:synth_graph_generator}.  


\begin{figure}
  \includegraphics[width=\columnwidth]{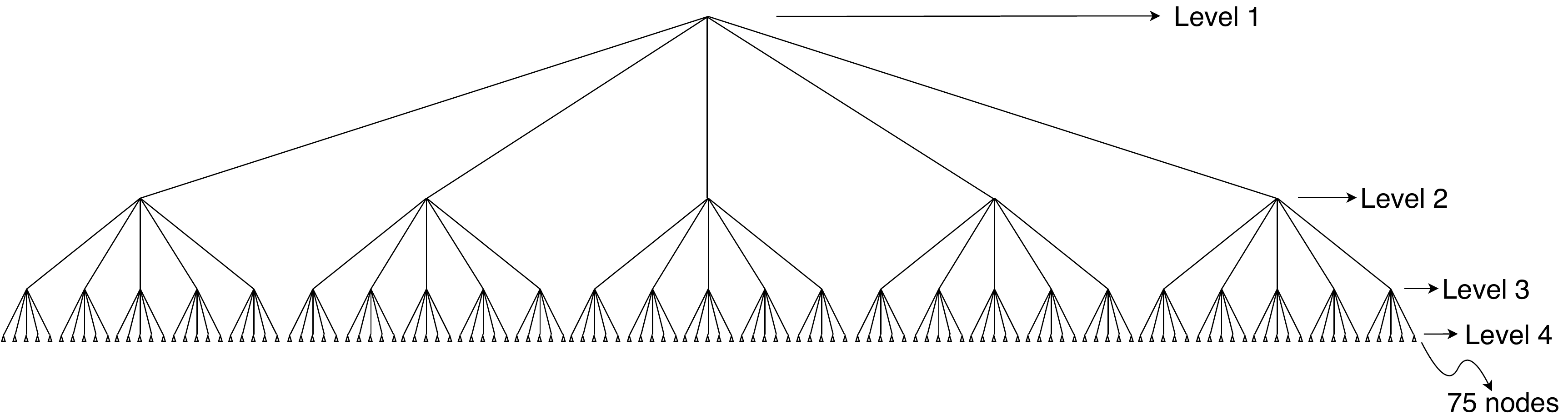}
  \caption{The hierarchical tree structure we use to generate our synthetic datasets. It is composed of $4$ levels, where the finest level has a branching factor of $75$, and all other levels have a branching factor of $5$.}
  \label{fig:hier}
\end{figure}

\label{ss_sec:synth_data}
In this experiment, we create a $5$-level hierarchical tree structure, whose leaves form the nodes in the graph. 
Each level in the hierarchical tree, except the level before the leaf nodes, which has a branching factor of $75$,
has a branching factor of $5$. 
Thus, the graph has $9375$ nodes. See Figure~\ref{fig:hier} for reference. We define the range of the \texttt{common-ratio} parameter between $\{1.05, 1.2, 1.4, 1.6, 1.8, 2.0\}$ (see Appendix~\ref{sec:synth_graph_generator} for details) . Higher values of the \texttt{common-ratio} results in fewer number of edges that are formed across nodes that appear in different communities. This results in progressively increasing the \emph{modularity} values of the graph as computed by the communities present in the second last level of the hierarchical community structure. On average, the \emph{modularity} value of the graph for the corresponding \texttt{common-ratio} is $0.23, 0.28, 0.33, 0.37, 0.41, 0.44$, respectively.
We use a power-law distribution to model the degree distribution of the graph, with the value $4.5$ for the power-law distribution parameter. The maximum degree a node has in the (directed) graphs we study is $187$ and the average degree is about $33$.

\subsection{Experimental Setup}

\begin{table}[t]
\small
\centering
\begin{threeparttable}
\caption{Real-world graph dataset statistics.}
\label{graph_stats}
\begin{tabular}{l@{\hspace{1em}}l@{\hspace{0.7em}}c@{\hspace{0.7em}}c@{\hspace{0.7em}}c@{\hspace{1em}}c@{\hspace{1em}}}
\toprule
  & & & & Initial & \\
  Dataset 	& 
  \#nodes 	& 
  \#edges 	& 
  \#labels  & 
  \#communities in & 
  Label  \\ 
  & & & & coarsened levels & rate \\   
\midrule
 \emph{BlogCatalog} & $10312$ & $667966$ & $34$ & $\{100, 10, 1\}$ & $0.17$ \\
 \emph{CS-CoAuthor} & $18333$ & $163788$ & $15$ & $\{135, 12, 1\}$ & $0.08$ \\
 \emph{DBLP}        & $20111$ & $115016$ & $23$ & $\{141, 11, 1\}$ & $2.86$ \\
\bottomrule
\end{tabular}
\footnotesize
Details of the graph datasets extracted from its largest connected component. \\ 
Label rate is the fraction of nodes in the training set. \\
The number of communities in each subsequent level is generated using $\sqrt{n}$, where, $n$ is the number of nodes in the graph in the current level. The last level in the hierarchy is created if the \#communities computes to be less than $10$, in which case we create the all-encompassing node. \\
The number of samples in the training set is chosen such that it results in the best performance in the non-hierarchical methods.
\end{threeparttable}
\vspace{-1em}
\end{table}

\subsubsection*{Setup of the Link Prediction Task}
We divide the original graph into three sets: validation set, test set and train graph. We sample edges (node pairs) such that the number of validation and test samples, considered as the unobserved set, equal $5\%$ and $10\%$ of the total number of edges, respectively. Further, we sample $99$ negative samples for each positive sample. 
%
We form the training graph using the set of edges in the train set, and we use this training graph to generate the representations for all the nodes. 
Then, for every edge in the validation and test sets, we compute the prediction score of the representations of its node pairs along with that of its corresponding negative samples and compute the mean average precision.

Moreover, to test our algorithm on link prediction using learnable decoders, we implement the \emph{DistMult} model~\cite{yang2014embedding} and a \emph{$2$-layer multi-layer perceptron} (MLP). We provide the element-wise product of the representations of the nodes that comprise an edge as input to train the above models. We use $2\%$ of the edges as the train set and $1\%$ each for the validation and test set, with $20$ negative samples for each positive edge, and report the average precision (AP) score of the test set for the best performing score on the validation set.

We run an elaborate hyper-parameter search, with
\texttt{context\_size}, \texttt{walk\_length}, and \texttt{walks\-}\texttt{\_per\_node} selecting values between  $\{2, 3, 4, 5\}$, $\{4, 6, 8, 10\}$, and from within $5$ and $60$, respectively, to generate the \emph{node2vec} representations.
The \texttt{p} and \texttt{q} parameters
takes values from sets $\{0.1, 0.25, 0.5, 0.75, 1, 2, 4, 6, 8\}$ and  
$\{1, 2, 4, 6, 8, 10\}$, respectively. The number of epochs is varied up to $4$. 
For \emph{HARP}, the $context\_size$ is chosen from $\{2, 3, 4, 6\}$, the $walk\_length$ from $\{5, 10, 20, 30, 40, 50\}$, and the $walks\_per\_node$ from $\{5, 10, 20, 30\}$. We choose $\beta$ and $\gamma$, hyper-parameters specific to the \emph{\mazi} model, from a more fine-tuned set for these graphs. $\beta$ and $\gamma$ is assigned values from $\{0.0, 0.25, 0.5, 1.0, 1.5, 1.25, 1.75, 2.0, 2.25, 2.5\}$ and $\{0.0, 1.0, 2.0,$ \\ $ 3.0\}$, respectively. Other hyper-parameters tuned in \emph{\mazi} include the number of epochs within an  optimization step in either direction, and the number of such optimization steps. These have been chosen such that they give the best performance for the respective datasets. In \emph{LouvainNE}, we use the stochastic node representations variant of their method which they use to report their best performing results. We perform a parameter sweep of the partitioning scheme provided by the approach for generating the hierarchy and also the damping parameter, which was given values such as $0.0001, 0.001, 0.01, 0.1, 1.0$. The number of dimensions for all methods have been set to $128$.

\subsubsection*{Setup of the Multi-label Classification Task}
We use a One-vs-Rest Logistic Regression model (implemented using LibLinear~\cite{fan2008liblinear}) with L2 regularization. 
For each graph dataset, we split the nodes into train, validation and test sets. 
In order to get a representative train set of the samples from each class, we sample a fixed number of instances, $s$, from each class. The validation and the test set is, thereafter, formed by almost equally splitting the remaining samples. In the case of the \emph{BlogCatalog} dataset, due to heavy class imbalance with respect to the number of instances in each class, we choose \texttt{min}($75\%$ of class samples, $s$) of samples in the train set. 
We choose the weight of the regularizer from the range \{$0.1$, $1.0$, $10.0$\},  such that it gives the best average macro F1 score on the validation set for the different methods. 
Overall, the number of samples in the training set is chosen such that it results in the best performance in the non-hierarchical methods, and then we reuse the same configuration for the hierarchical methods. We also perform a hyper-parameter search to find the best set of parameters that are specific to each method. 

To generate the best performing model of the approaches for evaluation, we conduct a search over the different hyper-parameters for the synthetic and the real-world graphs. For the synthetic graphs, \texttt{context\_size}, \texttt{walk\_length}, and \texttt{walks\_per\_node} in \emph{node2vec} are chosen from set $\{5, 10, 15\}$, $\{10, 20, 30\}$, and $\{10, 20, 30\}$, respectively. The \texttt{return} parameter, \texttt{p}, and the \texttt{in-out} parameter, \texttt{q}, takes on values between $\{0.25, 0.50, 1, 2, 4\}$ each. 
All but one graph gave the best performing model with number of epochs set to $1$, and thus, we limit the number of epochs to $2$. 
Node representations in \emph{HARP} and \emph{\mazi} are also generated using the above values for the \texttt{context\_size}, \texttt{walk\_length}, and  \texttt{walks\_per\_node} parameters.
Additionally, for the hyper-parameters specific to the \emph{\mazi} model, we choose both $\beta$ and $\gamma$ from the set $\{0.0, 1.0, 2.0\}$. Other hyper-parameters include the number of epochs within an optimization step in the fine to coarse direction and in the coarse to fine direction, and the number of such optimization steps. 
For the real world graphs, the \texttt{context\_size}, \texttt{walk\_length}, and \texttt{walks\_per\_node} parameters have been varied between $\{5, 10, 15\}$, $\{10, 20, 30, 40\}$, and $\{10, 20, 30, 40\}$. \texttt{p} and  \texttt{q} takes on values from the set $\{0.25, 0.50, 1, 2, 4\}$ each. 
\emph{LouvainNE}, we use the stochastic node representations variant of their method which they use to report their best performing results. We perform a parameter sweep of the partitioning scheme provided by the approach for generating the hierarchy and also the damping parameter, which was given values such as $0.0001, 0.001, 0.01, 0.1, 1.0$.
The number of dimensions for all methods have been set to $128$.

\begin{table}[t]
\centering
\begin{threeparttable}
\caption{Link prediction on real-world graphs.}
\label{link_pred}
\begin{tabular}{l@{\hspace{0.5em}}c@{\hspace{0.7em}}c@{\hspace{0.7em}}c@{\hspace{0.7em}}}
\toprule
  \multicolumn{1}{l}{} 	&
  \multicolumn{3}{c}{Mean Average Precision} \\	
  \multicolumn{1}{l}{Method} 	&
  \multicolumn{1}{c}{BlogCatalog} 	&
  \multicolumn{1}{c}{CS-CoAuth} 	&
  \multicolumn{1}{c}{DBLP} 	\\   
 \midrule
 \emph{node2vec}            & $0.534$ & $0.797$ & $0.914$ \\
 \emph{ComE}                & $0.389$ & $0.745$ & $0.896$ \\
 \emph{HARP w. $2$~lvls}	& $0.532$ & $0.755$ & $0.881$ \\
 \emph{HARP w. $3$~lvls}	& $0.460$ & $0.732$ & $0.874$ \\
 \emph{HARP w. all lvls}	& $0.126$ & $0.647$ & $0.769$ \\
 \emph{LouvainNE}           & $0.035$ & $0.270$ & $0.397$ \\
 \emph{\mazi}		        & $\textBF{0.587}$ & $\textBF{0.824}$ & $\textBF{0.930}$\\
\bottomrule
\end{tabular}
\scriptsize
Link prediction task performance of the methods is listed in the table. All \emph{HARP} variants use \emph{node2vec} as the base model. The mean average precision score is reported. The results are the average of $3$ runs. 
The standard deviation was observed to be less than $0.01$.
\end{threeparttable}
\end{table}

\begin{table}[t]
\centering
\begin{threeparttable}
\caption{Link prediction using learnable decoders on \emph{BlogCatalog}.}
\label{link_pred_mlp}
\begin{tabular}{l@{\hspace{0.7em}}c@{\hspace{0.7em}}c@{\hspace{0.7em}}c@{\hspace{0.7em}}}
\toprule
  \multicolumn{1}{l}{} 	&
  \multicolumn{1}{c}{Using} & 
  \multicolumn{1}{c}{} &
  \multicolumn{1}{c}{2-layer} \\	
  \multicolumn{1}{l}{Method} 	&
  \multicolumn{1}{c}{$\sigma$} & 
  \multicolumn{1}{c}{DistMult} &
  \multicolumn{1}{c}{MLP} \\	
 \midrule
 \emph{node2vec} & $0.62$ & $0.62$	& $0.59$ \\
 \emph{ComE} & $0.47$ & $0.47$	& $0.46$ \\
 \emph{HARP w. $2$~lvls} & $0.62$	& $0.62$ & $0.62$ \\
 \emph{HARP w. $3$~lvls} & $0.05$ & $0.56$	& $0.57$ \\
 \emph{HARP w. all lvls} & $0.05$	& $0.32$ & $0.43$ \\
 \emph{LouvainNE} & $0.07$ & $0.08$ & $0.12$ \\
 \emph{\mazi} & $\textBF{0.70}$	&  $\textBF{0.70}$	& $\textBF{0.69}$ \\
\bottomrule
\end{tabular}
\scriptsize
We report average precision score on link prediction task of the methods using \textbf{learnable decoders} - \emph{DistMult} and \emph{$2$-layer multi-layer perceptron}. ($\sigma$) is  short for the sigmoid function.
\end{threeparttable}
\end{table}

\begin{table}[t]
\small
\centering
\begin{threeparttable}
\caption{Multi-label node classification performance.}
\label{micro_macro}
\begin{tabular}{l@{\hspace{1.5em}}l@{\hspace{1.5em}}c@{\hspace{1.65em}}c@{\hspace{1.65em}}}
\toprule
  Method 	& 
  Dataset 	& 
  Micro F1 	& 
  Macro F1  \\ 
\midrule
 \emph{node2vec}	& & $0.3718~(0.00)$	& $0.2430~(0.00)$	\\
 \emph{ComE}        & & $\textBF{0.4016~(0.00)}$ & $0.2464~(0.00)$ \\
 \emph{HARP (n2v)}	& \emph{BlogCatalog} & $0.3602~(0.00)$ & $0.2418~(0.00)$ \\
 \emph{LouvainNE}   & & $0.2275~(0.00)$ & $0.1051~(0.00)$  \\ 
 \emph{\mazi}		& & $0.3874~(0.00)$	& $\textBF{0.2499~(0.00)}$	\\
 \midrule
 \emph{node2vec}	& & $0.8670~(0.00)$	& $0.8213~(0.00)$	\\
 \emph{ComE}        & & $0.8696~(0.00)$ & $0.8238~(0.00)$ \\
 \emph{HARP (n2v)}	& \emph{CS-CoAuth}  & $0.8634~(0.00)$   & $0.8153~(0.00)$	\\
 \emph{LouvainNE}   & & $0.7790~(0.00)$ & $0.7317~(0.00)$  \\ 
 \emph{\mazi}       & & $\textBF{0.8708~(0.00)}$   & $\textBF{0.8266~(0.00)}$   \\
\midrule
 \emph{node2vec}	& & $0.2499~(0.00)$ & $0.2314~(0.00)$    \\
 \emph{ComE}        & & $0.2517~(0.00)$	& $0.2323~(0.00)$ \\
 \emph{HARP (n2v)}  & \emph{DBLP} & $0.2515~(0.00)$ & $0.2326~(0.00)$   \\
 \emph{LouvainNE}   & & $\textBF{0.2578~(0.01)}$ & $\textBF{0.2367~(0.01)}$  \\ 
 \emph{\mazi} 	    & & $0.2510~(0.00)$	& 	$0.2317~(0.00)$ \\
\bottomrule
\end{tabular}
\footnotesize
Multi-label classification performance of \emph{node2vec}, \emph{HARP(n2v)} and \emph{\mazi} is listed on the table. The micro F1 and macro F1 scores are reported. 
For each method, we report the scores achieved on the test set such that it achieves the best macro F1 score in the validation set chosen from the relevant hyper-parameters associated with each method. The results are the average of three runs. The standard deviation up to $2$ decimal points is reported within the parentheses.
\end{threeparttable}
\vspace{-1em}
\end{table}

\begin{figure*}[h]
      \centering
      \subfloat[Micro F1 Score\label{fig:num_micro}]{%
        \includegraphics[width=0.4\textwidth]{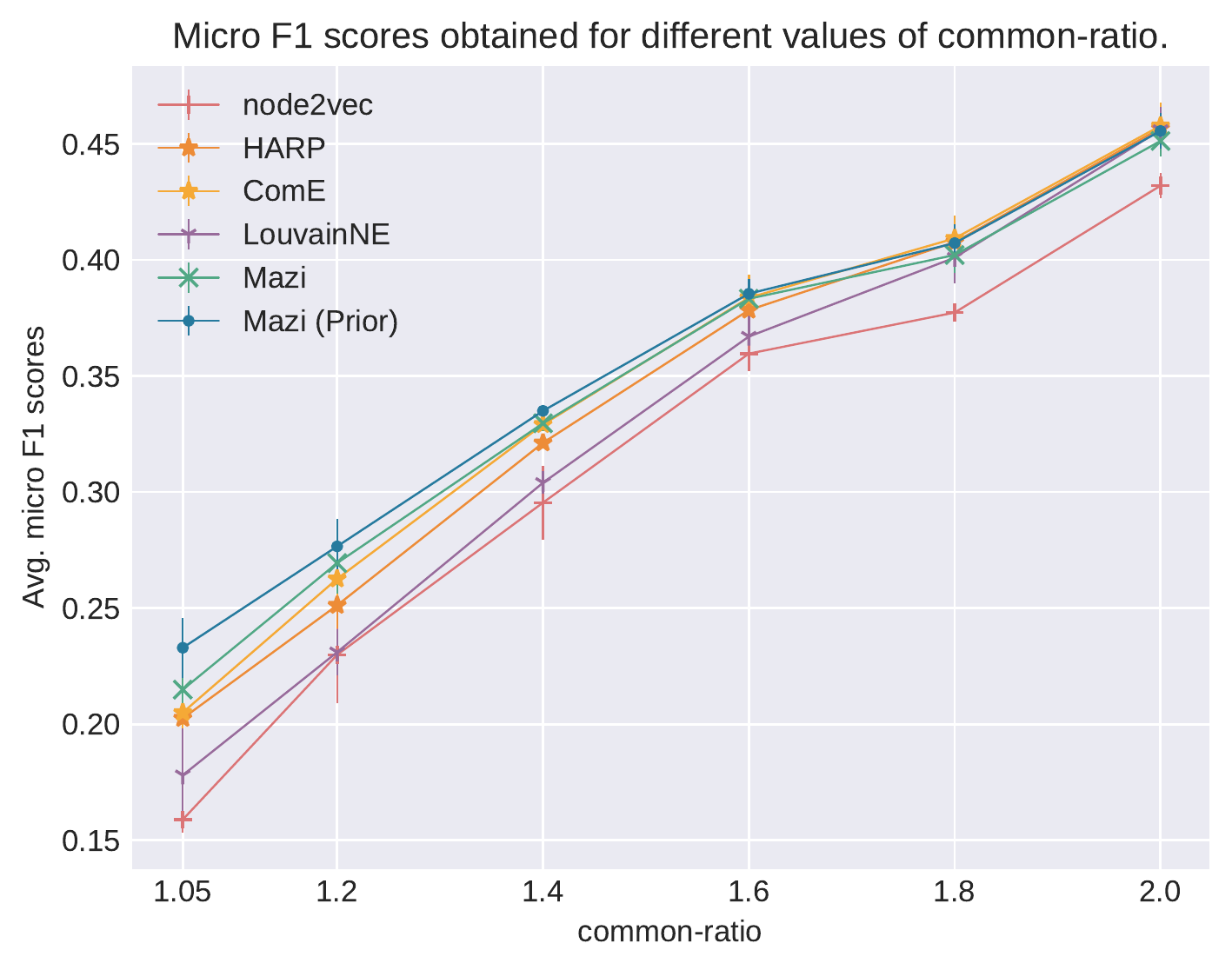}%
      }
      \hspace{20mm}
      \subfloat[Macro F1 Score\label{fig:num_macro}]{%
        \includegraphics[width=0.4\textwidth]{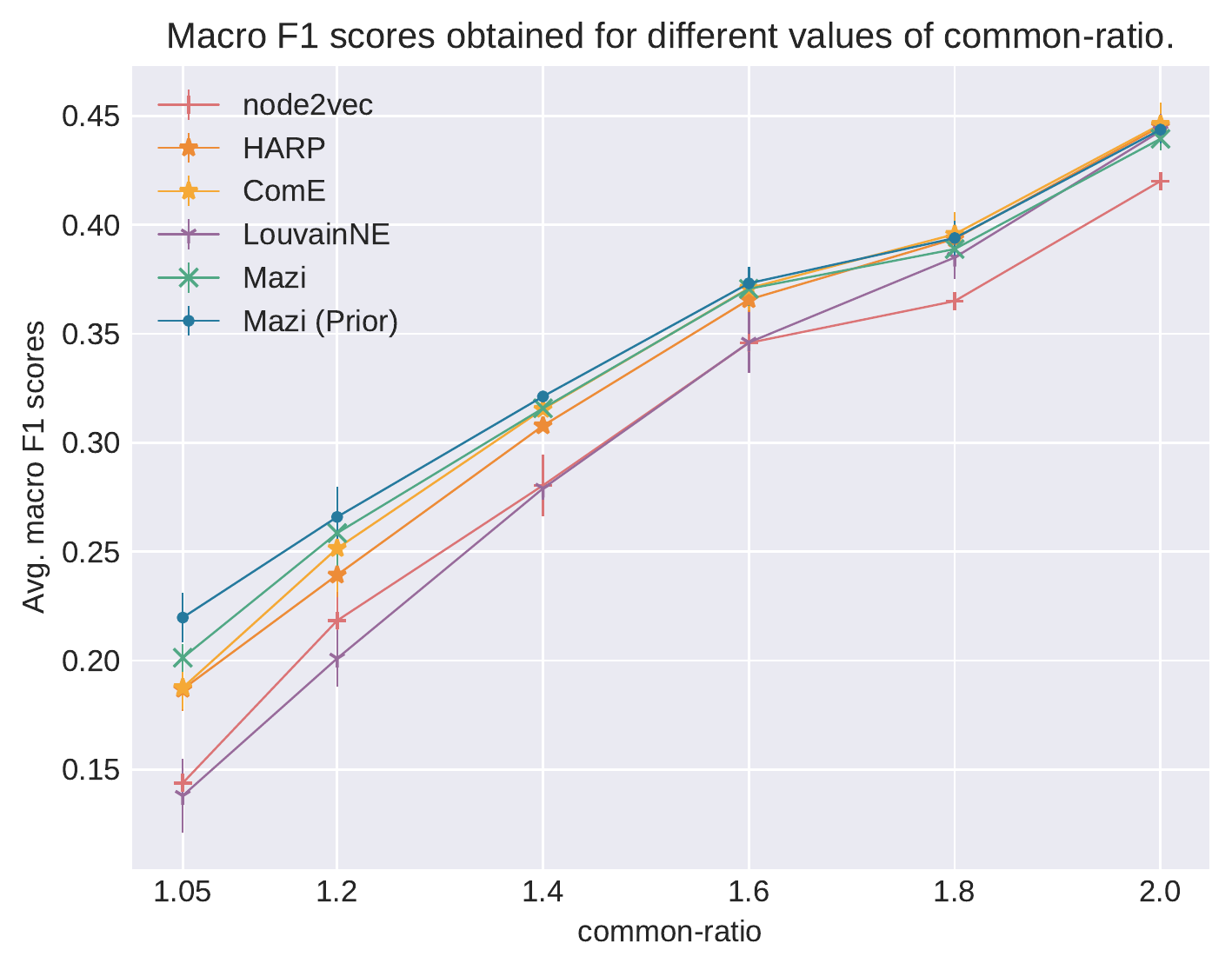}%
      }
      \caption{
      \footnotesize
      Average micro and macro F1 scores with standard deviation. Results are obtained using the methods: \emph{node2vec}, \emph{HARP}, \emph{LouvainNE}, \emph{ComE}, which \emph{uses} $\sqrt{n}$ communities, where, $n$ is the number of nodes in the graph, \emph{\mazi (Prior)}, which uses the community structure generated by the hierarchical clustering tree, and \emph{\mazi}, which \emph{generates} the community structure using Metis on the synthetic graphs with varying values of the parameter \texttt{common-ratio} generated 3 different times.}
    \label{fig:micro_macro_f1_scores}
\end{figure*}

\noindent
\subsection{Performance on the Link Prediction Task}
We evaluate \emph{Mazi} using the link prediction task on real world graph datasets.
\emph{\mazi} demonstrates good performance in the task over the competing approaches. The results are shown in Table~\ref{link_pred}. The gains observed in mean average precision (MAP) varies between $1.6\%$ in the \emph{DBLP} dataset to $10\%$ in the \emph{BlogCatalog} dataset over \emph{node2vec}. 
In comparison to \emph{HARP}, referred to as \emph{HARP w. all lvls} in Table~\ref{link_pred}, \emph{\mazi} shows gains as high as $366\%$ in \emph{BlogCatalog}. To study the performance of \emph{HARP}, 
we restrict the total number of levels to $2$, referred to as \emph{HARP w. 2 lvls}, and $3$, referred to as \emph{HARP w. 3 lvls}, and evaluate the performance of the method. We note that both these approaches result in better performance. Since \emph{HARP} chooses random edges and star-like structures to collapse, the coarsened graph in the last level formed by \emph{HARP} may not be indicative of the global structure of the network and further, not be indicative of how the edges actually form in the network. The negative samples can be, thus, scored relatively higher leading to low values of MAP. \emph{ComE}, using gaussian mixtures to model a single level of community representations, did not perform as well in the link prediction task. The best performing variant of \emph{LouvainNE}, as reported by the authors, uses random vectors for node representations for all nodes at every level in the hierarchy extracted out of the graph dataset. Since the node representation is created using a weighted aggregation of the different representations at every level in the hierarchy, \emph{LouvainNE} captures the hierarchical structure. However, it fails to capture the local neighborhood of a node such that nodes in close proximity are represented similarly. This may indicate the low performance of \emph{LouvainNE} on the link prediction task. 

In Table~\ref{link_pred_mlp}, we report the average precision (AP) scores using learnable models, \emph{DistMult} and a \emph{$2$-layer MLP}, on \emph{BlogCatalog}. We note very similar trends as in Table~\ref{link_pred} and observe that inspite of using learnable decoders, \emph{\mazi} outperforms all other approaches in this task. 

\noindent
\subsection{Performance on the Multi-label Classification Task}

\subsubsection*{Synthetic Graph Datasets.}
Figure~\ref{fig:micro_macro_f1_scores} plots the micro and the macro F1 scores on the multi-label node classification task on the synthetic datasets obtained by \emph{\mazi} using prior clustering, referred to as \emph{\mazi (Prior)}, \emph{\mazi} with the community structure initialized by Metis, referred to as \emph{\mazi (Metis)}, \emph{HARP}, \emph{LouvainNE}, \emph{ComE} and \emph{node2vec}. 
\emph{ComE} uses $\sqrt{n}$ communities for generating the node representations, where, $n$ is the number of nodes in the graph. \emph{HARP} uses \emph{node2vec} as its base model.
In \emph{\mazi (Metis)}, the hierarchical community structure is constructed using $4$ levels. The number of communities in
next coarser level is generated using
$\sqrt{n}$, where, $n$ is the number of nodes in the graph in the current level. The average gains observed in the macro F1 scores by \emph{\mazi (Prior)} against \emph{node2vec} range from over $50\%$ to $5\%$ for the \texttt{common-ratio} value of $1.05$ to $2.0$. Similar trends are observed in the micro F1 scores. 
As the \emph{modularity} of the graph, as defined by the finest level community structure, decreases, the random-walks in \emph{node2vec} will tend to stray outside the community and result in lowered performance. Since the labels are, however, distributed in accordance with the community structure of the graph, the objective in our method that minimizes the distance between the representation of a node to its community representation contributes to its improved performance. \emph{\mazi (Metis)} achieves similar performance as \emph{\mazi (Prior)} against \emph{node2vec}, ranging from $42\%$ to $5\%$ for \texttt{common-ration} $1.05$ to $2.0$.  

Further, \emph{\mazi (Prior)} and \emph{\mazi (Metis)} both are able to demonstrate significant benefits in comparison to \emph{HARP} for graphs with \texttt{common-ratio} ranging from $1.05$ to $1.6$. 
The average gain obtained by \emph{\mazi (Prior)} and \emph{\mazi (Metis)} in the macro F1 score are as high as $19\%$ and $9.5\%$, respectively, for the \texttt{common-ratio} $1.05$. 
We reason that for the graphs whose \emph{modularity}, 
as defined by the prior hierarchical community structure 
is low, the coarsening scheme of \emph{HARP}
is unable to capture a fitting hierarchical community structure
and thus, the representations learnt on the coarsest level 
does not result in good initializations for finer levels. 

\subsubsection*{Real World Graph Datasets.}
Table~\ref{micro_macro} reports the micro and macro F1 score obtained by all three methods on the real-world dataset. The datasets \emph{DBLP} and \emph{CS-CoAuthor} both exhibit high values of \emph{modularity}, that is, $0.83$ and $0.75$, respectively, while \emph{BlogCatalog} has a relatively lower \emph{modularity} value of $0.23$. In line with synthetic datasets, we note that \emph{\mazi} obtains a gain of up to $4.19\%$ and $7.55\%$ on macro F1 scores on \emph{BlogCatalog}, which has a lower \emph{modularity} value of $0.23$, against \emph{node2vec} and \emph{HARP} respectively.
We also note that \emph{ComE} obtains slightly better micro F1 score in \emph{BlogCatalog}. Its choice of using gaussian mixtures to model community distributions seems to capture the weak community structure in \emph{BlogCatalog} well. 
While the gain obtained in \emph{CS-CoAuthor} against \emph{node2vec} and \emph{HARP} is $0.43\%$ and $0.85\%$, respectively, in the macro F1 score, we observe that in \emph{DBLP}, whose \emph{modularity} value is the highest amongst the $3$ datasets, the performance of \emph{\mazi} is comparable with the competing approaches.

\input{sec-ablation}

%% file: sec-ablation.tex
\subsection{Ablation Study}
\label{sec:ablation}
We study the effect of the two parameters, $\gamma$ and $\beta$, that play an important role in determining the impact of the joint learning of the node representations and the hierarchical community structure on the node classification and the link prediction task. We set $\gamma$, which controls the 
contribution of the \emph{modularity} metric, $Q^l$ (Equation~\ref{modularity}), in 
the multi-objective function (Equation~\ref{eqn_dbl_x}), to $0$ to perform an ablation study on the same. 
Similarly, $\beta$, which determines the extent of the contribution of the proximity of a node representation to its community representation,
referred by $L^l_{comm}$ (Equation~\ref{comm_level})
in Equation~\ref{eqn_dbl_x}, is set to $0$ to perform an ablation study for that parameter.
Setting $\gamma = 0.0$ is equivalent to fixing the hierarchical community structure to its initial value and optimizing only the node representations while setting $\beta = 0.0$ is equivalent to fully ignoring the contribution of the proximity between the node and its community representations from the objective.

Table~\ref{ablation_mazi_macro_f1} depicts the performance of \emph{\mazi} on synthetic graphs generated using different  
values of the \texttt{common-ratio} for the node classification task, while Table~\ref{ablation_mazi_link_pred} depicts 
the performance of the method on the link prediction task. For the node classification task, we compare the above models using (i) \emph{\mazi} with prior community structure generated by the hierarchical solution, referred to as \emph{\mazi (Prior)}, and (ii) \emph{\mazi} using $4$ levels in the hierarchical community structure. The initial community structure is generated by \emph{Metis}, where the number of clusters is equal to the square-root of the number of nodes in the previous level finer graph. 

To study the effectiveness of \emph{\mazi} on the link prediction task, we study the performance of \emph{\mazi} using the initial community structure generated by \emph{Metis}, where the number of clusters is determined similar to above, and compare the scores obtained with \emph{\mazi} without using $Q^l$, obtained by setting $\gamma=0.0$, and \emph{\mazi} without using $L^l_{comm}$, obtained by setting $\beta=0.0$.

\paragraph{Performance of \emph{\mazi} (Prior) on the node classification task.}
A non-zero value of $\beta$ plays a crucial role in extracting good performance of \emph{\mazi} using the prior community structure. Since the representations learned are benefited by the knowledge of a fitting community structure, performance achieved by $\beta=0.0$ is consistently lower than when $\beta \neq 0.0$. We also note that in many of these graphs, a non-zero $\gamma$ value does not contribute to the best performance. Since the graph has been generated using the prior community structure, which is also used by the synthetic label generating procedure, refining it further has not resulted in better performance.  

\paragraph{Performance of \emph{\mazi} on the node classification task.}
The effect of $\gamma$ is more apparent in \emph{\mazi} using the \emph{Metis} community structure. Since the community structure provided by \emph{Metis} does not fully conform to the prior community structure and the label distribution on the synthetic graphs correlate with the finest level community structure, we note that refining the hierarchical community structure and thereby, using it to improve the representations lead to better performance of the model. 

\paragraph{Performance of \emph{\mazi} on the link prediction task.}
For the real-world datasets, we report effectiveness of $\gamma$ and $\beta$ in Table~\ref{ablation_mazi_link_pred}. 
In all the real-world datasets, we note that the datasets achieve better performance when accounting for non-zero values of the $\beta$.
This is especially evident in the \emph{BlogCatalog} dataset, wherein \emph{\mazi} shows a gain as high as $4.09\%$ when compared to \emph{\mazi} which sets $\beta$ to $0.0$. Further, we observe that the community structure refinement in \emph{BlogCatalog} and \emph{CS\_CoAuthor}, when $\gamma \neq 0.0$, leads to better performance, whereas in \emph{DBLP}, the results obtained are comparable to when we do not account for refinement in the community structure.

\begin{table}[t]
\small
\centering
\begin{threeparttable}
\caption{Ablation study on the synthetic datasets for the node classification task.}
\label{ablation_mazi_macro_f1}
\begin{tabular}{@{\hspace{0.5em}}l@{\hspace{0.1em}}c@{\hspace{0.5em}}r@{\hspace{0.5em}}r@{\hspace{0.5em}}r@{\hspace{0.5em}}r@{\hspace{0.5em}}r@{\hspace{0.5em}}}
\toprule
  Method  &
  common  &	
  $\gamma=0.0$ & 
  \multicolumn{1}{c}{\% gain} &
  $\beta=0.0$ &  
  \multicolumn{1}{c}{\% gain} &
  $\gamma\neq0.0$ \\ 
  &
  ratio &
   	    & 
  \multicolumn{1}{c}{w/o $Q^l$} &
        & 
  \multicolumn{1}{c}{w/o $L_{comm}^l$} &
  $\beta\neq0.0$ \\ 
\midrule
\emph{\mazi (Prior)} & 1.2   & \textBF{0.2671}    & -0.415 (0.607) & 0.2551 & 4.256 (0.555)  & 0.2659 \\
                    & 1.4   & 0.3210    & 0.073 (0.101) & 0.3155 & 1.830 (1.118)   & \textBF{0.3213}  \\
                    & 1.6   & \textBF{0.3735}    & -0.098 (0.128) & 0.3690   & 1.140 (0.099)  & 0.3732 \\
                    & 1.8   & 0.3936 & 0.093 (0.118)  & 0.3870 & 1.774 (0.982)  & \textBF{0.3939} \\
                    & 2.0   & 0.4437 & 0.000 (0.000)   & 0.4372  & 1.482 (0.628)   & \textBF{0.4437} \\
\midrule
\emph{\mazi}         & 1.2   & 0.2578   &  0.283 (0.889)  & 0.2561  & 0.970 (0.908)  & \textBF{0.2585} \\
                    & 1.4   & 0.3140    & 0.626 (0.035)    & 0.3142 & 0.690 (0.079)   & \textBF{0.3164} \\
                    & 1.6   & 0.3705  & 0.033 (0.546)   & 0.3688 & 0.492 (1.005)     & \textBF{0.3706} \\
                    & 1.8   & 0.3878  & 0.270 (0.414)  & 0.3873  & 0.390 (0.579)    & \textBF{0.3889} \\
                    & 2.0   & 0.4386 & 0.199 (0.328)    & 0.4380  & 0.350 (0.022)  & \textBF{0.4395} \\
\bottomrule
\end{tabular}
\footnotesize
\begin{tablenotes}
\item The macro F1 scores and the percent gain achieved by \emph{\mazi} over \emph{\mazi} without $Q^l$ (from Equation~\ref{modularity}) by setting $\gamma=0.0$ and \emph{\mazi} without $L^l_{comm}$ from (Equation~\ref{comm_level}) by setting $\beta=0.0$
are reported for the graphs synthetically generated in $3$ runs.
$\beta$ controls weight of the contribution of the similarity between the representations of a node to its community in the next coarser level in the multi-objective function. $\gamma$ controls the weight of the contribution of the \emph{modularity} metric in the multi-objective function. Hyper-parameters controlling the structure of the synthetic graphs are detailed in Section~\ref{ss_sec:synth_data}. The standard deviation up to $3$ decimal points is reported within the parentheses.
\end{tablenotes}
\end{threeparttable}
\vspace{-1em}
\end{table}

\begin{table}[t]
\small
\begin{threeparttable}
\caption{Ablation study for link prediction.}
\label{ablation_mazi_link_pred}
\begin{tabular}{@{\hspace{0.1em}}l@{\hspace{0.1em}}c@{\hspace{0.1em}}c@{\hspace{0.1em}}c@{\hspace{0.1em}}c@{\hspace{0.1em}}c@{\hspace{0.1em}}}
\toprule
  Graph	& 
  $\gamma=0.0$ & 
  \multicolumn{1}{c}{\% gain} &   
  $\beta=0.0$ &  
  \multicolumn{1}{c}{\% gain} & 
  $\{\gamma, \beta\} \neq0.0$ \\ 
        &
        &
  \multicolumn{1}{c}{without $Q^l$} &
        &
  \multicolumn{1}{c}{without $Q^l$} &
         \\
  \midrule  
  \emph{BlogCatalog}  & $0.5862$ & 0.154 (0.034)& $0.5641$ & 4.089 (0.087) &$\textBF{0.5871}$ \\
  \emph{CS\_CoAuth} & $0.8234$ & 0.028 (0.122) & $0.8212$ & 0.292 (0.088) & $\textBF{0.8236}$ \\
  \emph{DBLP}         & $\textBF{0.9301}$ & -0.021 (0.077) & $0.9292$ & 0.075 (0.172) & $0.9299$ \\  
\bottomrule
\end{tabular}
\footnotesize
\begin{tablenotes}
\item The mean average precision scores and the corresponding percent gain, averaged over $3$ runs, achieved by \emph{\mazi} on the link prediction task over \emph{\mazi} without $Q^l$ (from Equation~\ref{modularity}) by setting $\gamma=0.0$ and \emph{\mazi} without $L^l_{comm}$ from (Equation~\ref{comm_level}) by setting $\beta=0.0$ is reported for the real-world graphs.
$\beta$ controls the weight of the contribution of the proximity of a representation of a node to its community representation in the subsequent level in the multi-objective function. 
$\gamma$ controls the weight of the contribution of the \emph{modularity} metric in the multi-objective function.
\end{tablenotes}
\end{threeparttable}
\vspace{-1em}
\end{table}

%% file: sec-related.tex
\section{Related Work}
\label{sec:related}

\paragraph{Graph Representation Learning}
Several methods model node representations using deep learning losses in supervised, semi-supervised and unsupervised settings. Amongst the unsupervised methods, the Skip-gram model is a popular approach used in the literature~\cite{perozzi2014deepwalk, grover2016node2vec, tang2015line} to model the local neighborhood of a node using random walks while learning its representation. However, unlike our method, these representations are inherently flat and do not account for the hierarchical community structure that is present in the network.

\paragraph{Community-aware representation learning}
Existing methods have also explored jointly learning communities at a single level and the representations of the nodes in the graph~\cite{cavallari2017learning, sun2019vgraph}. \emph{ComE}~\cite{cavallari2017learning}
models the community and the node representations using a gaussian mixture formulation. \emph{vGraph}~\cite{sun2019vgraph} assumes each node to belong to multiple communities and a community to contain multiple nodes, and parametrizes the node-community distributions using the representations of the nodes and communities. Unlike these approaches, our approach utilizes the inductive bias introduced by the hierarchical community structure in the representations. 

\paragraph{Hierarchical Representation Learning}
Recently, unsupervised hierarchical representation learning methods have been explored to leverage the multiple levels that are formed by hierarchical community structure in the graph.
\emph{HARP}~\cite{chen2018harp} and \emph{LouvainNE}~\cite{bhowmick2020louvainne} both learn the node representations of a graph by utilizing a hierarchical community structure. \emph{HARP} uses an existing node representation learning method, such as \emph{node2vec}, to generate node representations for graphs at coarser levels and use them as initializations for learning the representations of the nodes at finer levels. 
\emph{LouvainNE} recursively generates sub-communities within each community for a graph. The representations for a node in all the different sub-communities are generated either stochastically or using one of existing flat representation learning method, which is then subsequently aggregated in a weighted fashion to form the final node representation.    
\emph{SpaceNE}~\cite{long2019hierarchicalsubspace} constructs sub-spaces within the feature space to represent the hierarchical community structure, and learns node representations that preserves proximity between vertices as well as similarities within communities and across communities.
However, all these approaches consider a static hierarchical community structure, which is then utilized to influence the node representations. In comparison, we jointly learn the node representations and the hierarchical community structure that is influenced by the node representations.

In a parallel line, some GNN-based methods have been suggested to model the hierarchical structure present in the graph while learning the network representations. Some of these methods generate representations for the entire graph~\cite{ying2018hierarchical,huang2019attpool} and are useful for the graph classification task. 
For the node representation learning task, a recent approach includes \emph{HC-GNN}~\cite{zhong2020hierarchical}.
\emph{HC-GNN} uses the representation of a node's community at each level in the aggregation and combine phase of the GNN framework.
\emph{GXN}~\cite{li2020graph}, another GNN model, introduces a pooling method along with a novel idea of feature crossing layer which allows feature exchange across levels. 
However, these are supervised methods and use task specific losses while considering static hierarchical community structures.

%% file: sec-conclusion.tex
\vspace{-1mm}
\section{Conclusion}
\label{sec:con}

This paper develops a novel framework, \emph{\mazi}, for joint unsupervised learning
of node representations and the hierarchical community structure in a given graph.
At each level of the hierarchical structure, \emph{\mazi} coarsens the graph and learns the node representations, and 
leverages them to discover communities in the hierarchical structure. 
In turn, \emph{\mazi} uses the structure to learn the representations. 
Experiments conducted on synthetic and real-world graph datasets in the node classification and link prediction demonstrate the competitive performance of the learned node representations compared to competing approaches.

%% file: sec-appendix.tex
\section{Supplementary Material}
\label{sec:appendix}

\subsection{Synthetic Graph Generator}
\label{sec:synth_graph_generator}

\paragraph{Synthetic Graph Generator Model} Our model generates a graph respecting
a hierarchical community structure by modeling this structure using a hierarchical tree.
Each level in the hierarchical tree corresponds to a level in the hierarchical community structure of the generated graph.
The nodes at each level of the tree structure forms the  communities at that level in the hierarchical community structure. 
The nodes in the last level of the hierarchical tree structure, or, the leaves of the tree, forms the nodes of the generated graph.
Further, we also ensure that the generated graph emulates the characteristics of real-world networks. 
First, the nodes are constructed such that a node in the graph is, in expectation, able to form edges with other nodes in communities associated with upper levels in the hierarchical community structure. 
Typically, the number of edges a node forms with nodes in other communities at upper levels progressively decrease as we go up the hierarchy. To achieve this, we model the expected number of edges using a probability distribution generated from a geometric progression. A geometric progression is a series of numbers where each number after the first is the product of the preceding term with a constant, non-one number called the common ratio. Thus, we accept a parameter, referred to as \texttt{common-ratio}, to enable us to compute $L$ terms in the series, one corresponding to each level in the $L$-level hierarchical community structure. Using these $L$ terms, we compute the probability distribution of a node to form an edge with another in a community present in different levels in the hierarchy. This parameter plays a key role in determining the \emph{modularity} of the generated graph using the communities formed by the hierarchical structure. Further, the degrees associated with the nodes in the graph use a power distribution to model the behavior of real-world networks. Other properties that we tune are the maximum degree of a node, number of levels in the hierarchical tree structure, branching factor of nodes in the intermediate levels, the number of leaves, among others. 
\paragraph{Synthetic Label Generation Procedure} To aid us in the node classification task, we generate labels for the nodes such that they correlate with the hierarchical structure of the graph. For each node, we create a probability distribution over the unique communities present in the second last level of the hierarchy. The weight corresponding to each community in this probability distribution is determined by the frequency of nodes in that community the node is connected to. Using this probability distribution, we sample a community id which serves as its label. The total number of labels is, thus, equal to the number of communities in the second last level in the hierarchical structure.
